%% file: main.tex
\documentclass[runningheads]{llncs}

\usepackage{eccv}

\usepackage{eccvabbrv}

\usepackage{graphicx}
\usepackage{booktabs}

\usepackage[accsupp]{axessibility}  

\input{Configs/00_packages}
\input{Configs/01_definations}

\usepackage{hyperref}

\usepackage{orcidlink}

\begin{document}

\title{Reconstruction-Anchored Diffusion Model for Text-to-Motion Generation} 


\author{Yifei Liu\inst{1,2} \and
Changxing Ding \inst{1} \thanks{Corresponding author.} \and
Ling Guo \inst{1} \and
Huaiguang Jiang \inst{1} \and \\
Qiong Cao \inst{2} }

\authorrunning{Y.~Liu et al.}

\institute{South China University of Technology \\
\email{\{ft\_lyf, 202510182670\}@mail.scut.edu.cn, \\ 
\{chxding, hihuagong2021\}@scut.edu.cn}
\and Joy Future Academy \\
\email{mathqiong2012@gmail.com}
}
\maketitle

\input{Sec/Sec_00_abstract}

\input{Sec/Sec_01_introduction}
\input{Sec/Sec_02_related_work}
\input{Sec/Sec_03_method}
\input{Sec/Sec_04_experiments}
\input{Sec/Sec_06_conclusion}



%
%
\bibliographystyle{splncs04}
\bibliography{Bibs/sample-base}

\clearpage
\appendix
\input{Sec/Sec_07_sup}

\end{document}

%% file: Configs/00_packages.tex
\usepackage{algorithm}
\usepackage{algorithmic}
\usepackage{booktabs} %
\usepackage{adjustbox}
\usepackage{amsmath} 
\usepackage{multirow}
\usepackage[table,dvipsnames]{xcolor}
\usepackage{xspace}
\usepackage{url}
\usepackage{arydshln}
\usepackage{amssymb}
\usepackage{colortbl}

\usepackage{booktabs}
\usepackage{tabularx}


%% file: Configs/01_definations.tex
\newcommand{\aref}[1]{Appendix~\ref*{#1}}

\newcommand{\highlight}[1]{#1}

\newcommand{\modelname}{{RAM}\xspace}
\newcommand{\modelnamefull}{{\textbf{R}econstruction-\textbf{A}nchored Diffusion \textbf{M}odel}\xspace}
\newcommand{\modelnamefullrelax}{{Reconstruction-Anchored Diffusion Model}\xspace}

\definecolor{DeltaColor}{rgb}{0.039,0.73,0.71}
\definecolor{SigmaColor}{rgb}{0.98,0.45,0.0}
\definecolor{AlphaColor}{rgb}{0,0,0.8}
\definecolor{BetaColor}{rgb}{0.8,0,0.8}
\definecolor{GammaColor}{rgb}{0.514,0.34,0.224}
\definecolor{EpsilonColor}{rgb}{0.353,0.725,0.906}
\definecolor{PurpleColor}{rgb}{0.5,0,0.7}
\definecolor{OrangeColor}{rgb}{0.914,0.541,0.141}
\definecolor{GreenColor}{rgb}{0.137,0.573,0.565}
\definecolor{RedColor}{rgb}{0.949,0.275, 0.224}
\definecolor{LightCyan}{rgb}{0.88,1,1}
\definecolor{Gray}{gray}{0.3}
\definecolor{Strawberry}{rgb}{1,0.26,0.64}

\definecolor{BetaColor}{rgb}{0.8,0,0.8}

\definecolor{LightCyan}{rgb}{0.88,1,1}
\definecolor{lightgray}{rgb}{0.9,0.9,0.9}

\newcommand{\qheading}[1]{\noindent\textbf{#1}}

\definecolor{GreenColor}{rgb}{0.137,0.573,0.565}

\renewcommand{\paragraph}[1]{\medskip\noindent\textbf{#1}\ \ }

\makeatletter
\newcommand*\bigcdot{\mathpalette\bigcdot@{.5}}
\newcommand*\bigcdot@[2]{\mathbin{\vcenter{\hbox{\scalebox{#2}{$\m@th#1\bullet$}}}}}
\makeatother

%% file: Sec/Sec_00_abstract.tex
\begin{abstract}
Diffusion models have seen widespread adoption for text-driven human motion generation and related tasks due to their impressive generative capabilities and flexibility. However, current motion diffusion models face two major limitations: a representational gap caused by pre-trained text encoders that lack motion-specific information, and error propagation during the iterative denoising process. This paper introduces \modelnamefull (\textbf{\modelname}) to address these challenges. First, \modelname leverages a motion latent space as intermediate supervision for text-to-motion generation. To this end, \modelname co-trains a motion reconstruction branch with two key objective functions: self-regularization to enhance the discrimination of the motion space and motion-centric latent alignment to enable accurate mapping from text to the motion latent space. Second, we propose Reconstructive Error Guidance (REG), a testing-stage guidance mechanism that exploits the \highlight{motion} diffusion model's inherent self-correction ability to mitigate error propagation. At each denoising step, REG uses the motion reconstruction branch to reconstruct the previous estimate, reproducing the prior error patterns. By amplifying the residual between the current prediction and the reconstructed estimate, REG highlights the improvements in the current prediction. Extensive experiments demonstrate that \modelname achieves significant improvements and state-of-the-art performance. Our code will be released on \url{https://feifeifeiliu.github.io/RAM} .
  \keywords{Human Motion Generation \and Text-to-Motion Generation \and Diffusion Models \and Representation Learning \and Diffusion Guidance}
\end{abstract}

%% file: Sec/Sec_01_introduction.tex
\section{Introduction}

\input{Fig/Fig_00_teaser}

Imagine giving a textual description and immediately witnessing a lifelike avatar execute it, with physically plausible, faithful body movements in the correct sequence. This vision drives human motion generation with applications in virtual reality \cite{du2023avatars}, game content creation \cite{liang2024omg}, and embodied robotics \cite{xia2021relmogen}. The task is inherently challenging because the language is abstract. At the same time, motion is continuous, high-dimensional, and kinematically constrained—demanding both fine-grained semantic understanding and robust many-to-many mappings between natural language and human motion dynamics.

This challenge has sparked extensive research interest, which can be broadly categorized into two main approaches: VQ-VAE-based and diffusion-based methods. Among these, diffusion-based methods have gained widespread adoption across downstream tasks, including motion in-betweening \cite{cohan2024flexible}, human-object interaction \cite{li2024controllable}, and multi-human motion modeling \cite{liang2024intergen, sun2025beyond, wang2025timotion, Dai2026TCDiff++}, owing to their exceptional flexibility and controllability. Existing diffusion-based methods typically leverage pre-trained text encoders to obtain robust textual embeddings, such as T5 \cite{ni2021sentence}, CLIP \cite{radford2021learning}, and DistilBERT \cite{sanh2019distilbert}. Conditioned on these textual embeddings, motion diffusion models learn to recover motion data from noise via iterative denoising. Recent advances have incorporated various techniques, including latent diffusion \cite{chen2023executing}, preference optimization \cite{sheng2024exploring}, hierarchical semantic graphs \cite{jin2023graphmotion}, and retrieval-augmented generation \cite{zhang2023remodiffuse}, achieving notable improvements in inference speed, motion realism, and semantic-motion alignment.

Nevertheless, motion diffusion models still face severe limitations in both text models and the denoising process. First, pre-trained text models typically lack cues about motion dynamics. Although CLIP captures visual concepts that correlate with actions, it fails to encode essential temporal dynamics and kinematic constraints, as it was trained exclusively on static image-text pairs. This absence forces models to bridge an unnecessarily large representational gap, hindering the learning of accurate semantic-to-dynamic mappings. Second, diffusion models suffer from error propagation \cite{li2023error}. More specifically, early denoising steps, which must recover motion from nearly pure noise, are particularly prone to generating error patterns. Once such artifacts emerge, they can propagate across subsequent denoising steps, leading to a degradation in sample quality.

Herein, we introduce \modelnamefullrelax (\textbf{\modelname}), a novel diffusion-based framework to address these challenges. For the first problem, we leverage the latent space learned from motion reconstruction as an intermediate supervisory signal for text-to-motion generation. Specifically, \modelname employs a two-stream pipeline \cite{ahuja2019language2pose}: motion reconstruction—where the diffusion model reconstructs motion sequences conditioned on motion-encoder latents; and text-to-motion generation—where the same diffusion model generates motion from text-encoder latents. Based on this pipeline, \modelname incorporates two objectives: (a) \textbf{self-regularization}, which computes a cross-entropy loss in the motion latent space to enhance discrimination between motion latents, helping to learn a compact yet expressive motion representation; and (b) \textbf{motion-centric latent alignment}, aligning the text latent space with the motion latent space, with carefully designed gradients to ensure stable end-to-end training. Together, these designs enable \modelname to map text embeddings into a latent space sensitive to motion, inherently embedding the dynamic features required for realistic motion synthesis and bridging the representation gap.

To address the second problem, we introduce the Reconstructive Error Guidance (\textbf{REG}), which harnesses the self-correction capabilities of \highlight{text-to-motion} diffusion models to mitigate error propagation. Our core insight is that diffusion models can inherently self-correct, as they do when restoring clean data from noise. To maximize this property, at each denoising step during the testing stage, the motion reconstruction branch reconstructs the previous estimate, thereby capturing earlier error patterns. We then calculate the residual between the current text-driven prediction and the reconstruction, and integrate it into the prediction to generate the final output. This residual highlights the improvements in the current prediction. By amplifying this term, REG directs the sampling process away from error-prone regions, thereby reducing error propagation and enhancing the quality of generated motions throughout denoising.

By integrating these core innovations, \modelname enables the generation of more realistic and semantically aligned motions from text. Experiments show that \modelname achieves
state-of-the-art performance: on the HumanML3D dataset \cite{guo2022humanml3d}, \modelname achieves an R-Precision@1 of 56.1\% and an FID of 0.032 with only 20 inference steps—surpassing previous diffusion-based methods and outperforming most VQ-VAE-based models, which have historically been stronger than diffusion-based ones on this metric. Consistent performance gains are also observed on the KIT-ML dataset \cite{plappert2016kit}. Comprehensive ablation studies further confirm that each component makes a meaningful contribution to overall performance improvements.

%% file: Fig/Fig_00_teaser.tex
\begin{figure}
    \centering
    \includegraphics[width=1.0\linewidth]{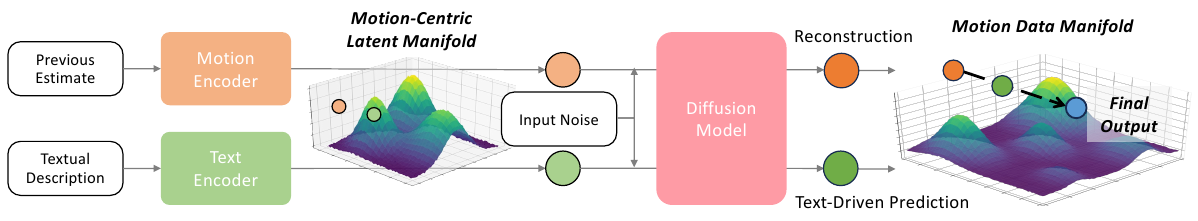}
    \caption{At inference time, \modelname first maps a textual description onto a motion-centric latent manifold and then predicts using a diffusion model. Meanwhile, it reconstructs previous estimates that contain error patterns. By contrasting these predictions, \modelname uses the reconstruction as a negative reference to drive the output away from poor estimates and towards the real data manifold. Best viewed in color.
    }
    \label{fig:teaser}
\end{figure}

%% file: Sec/Sec_02_related_work.tex
\nopagebreak

\section{Related Work}

\paragraph{Text-Driven Human Motion Generation.}Current research on text-to-motion generation has consolidated mainly around two principal families: diffusion models and vector-quantized variational autoencoders (VQ-VAE). Early diffusion-based approaches such as Motion Diffusion Model \cite{tevet2022mdm} and MotionDiffuse \cite{zhang2022motiondiffuse} trained denoising networks directly in the raw motion space, followed by a series of extensions that target finer semantic alignment \cite{zhang2023finemogen}, open-vocabulary coverage \cite{liang2024omg}, retrieval-enhanced consistency \cite{zhang2023remodiffuse}, keyframe-centric stability \cite{bae2025less}; recent advances further include step-aware fine-tuning \cite{tan2026easytune}, part-level motion composition \cite{li2026frankenmotion}, and reward-guided alignment \cite{weng2026realign}. In parallel, latent diffusion methods first encode motions into a continuous latent space and denoise them there, aiming to improve efficiency and quality, \eg, MLD \cite{chen2023executing}, MotionLCM \cite{dai2024motionlcm}, Salad \cite{hong2025salad}. In contrast, VQ-VAE-based pipelines—pioneered by T2M-GPT \cite{zhang2023t2mgpt} and advanced through MMM \cite{pinyoanuntapong2024mmm}, MoMask \cite{guo2024momask}, BAMM \cite{pinyoanuntapong2024bamm}, MoGenTS \cite{yuan2024mogents}, BAD \cite{hosseyni2025bad}, KinMo \cite{zhang2025kinmo}, and LaMP \cite{li2025lamp}—have empirically exhibited higher motion fidelity, typically reflected in lower FID scores than their diffusion counterparts. 
In this paper, \modelname demonstrates that \textit{diffusion-based approaches can achieve FID performance comparable to that of VQ-based approaches}.

\paragraph{Pre-trained Text Models and Two-Stream Methods.}Since text-to-motion datasets are significantly smaller than typical text or text–image datasets, most methods utilize pre-trained text models to extract robust text embeddings. CLIP is widely used for its visual-textual embedding space \cite{tevet2022motionclip}, but recent research suggests that it may not be optimal for aligning text and motion. Instead, these studies suggest fine-tuning text encoders to explicitly learn a joint language–motion embedding space \cite{maldonado2025moclip, zhang2025kinmo}. This approach can be traced back to early two-stream methods \cite{ahuja2019language2pose}, which use dual branches—motion reconstruction and text-to-motion generation—and share a decoder to learn a joint language-motion space implicitly. Subsequent works further constrain this space using latent alignment, KL divergence, or contrastive learning \cite{ghosh2021synthesis, petrovich2022temos, petrovich2023tmr}. These approaches inspire our method but differ fundamentally: we center the alignment on a carefully designed motion latent space, with the text space aligning to it. We demonstrate that, when employing a diffusion model as the decoder, focusing on modeling detailed motion dynamics yields better motion synthesis results than forcing the learning of a joint language-motion space.

\paragraph{Diffusion Guidance.}Guidance in diffusion sampling typically combines multiple score estimateps to enrich the effective target distribution or to impose auxiliary conditioning \cite{dhariwal2021diffusion, ho2022classifier, karras2024guiding}. Common estimates include conditional score 
estimates $\nabla_{x_t} \log p(x_t | t, c)$, unconditional score estimates $\nabla_{x_t} \log p(x_t | t)$, classifier gradients $\nabla_{x_t} \log p(y | x_t)$, and CLIP-derived similarity gradients \cite{dhariwal2021diffusion, nichol2021glide, ho2022classifier}. Recent works further introduce deliberately weakened
scores by degrading the predictor—\eg, applying dropout \cite{karras2024guiding}, skipping layers \cite{stabilityai_sd3.5_2024}, or perturbing attention \cite{ahn2024self}. These weak scores function as contrastive references: amplifying samples favored by stronger scores while suppressing those aligned with weaker ones improves fidelity and semantic alignment. In the same spirit, we derive a weakened
score by conditioning the predictor on a motion latent that carries previously introduced error patterns, and use it as a contrastive reference within our guidance mechanism.

%% file: Sec/Sec_03_method.tex
\section{Method} 

\paragraph{Overview.}\modelname generates a sequence of realistic human motion from a given text description. This process starts by extracting text embeddings with pre-trained text models. These embeddings are mapped onto a motion latent manifold and decoded into a motion sequence using a diffusion model. \modelname also reconstructs the past motion estimate as a negative reference in the inference stage, as illustrated in \cref{fig:teaser}. By guiding predictions away from this reference, \modelname achieves improved sampling quality. 

For a thorough understanding of \modelname, we begin by presenting the overall architecture, which features two main branches: motion reconstruction and text-to-motion generation (\cref{sec:arch}). Next, we detail the training objectives, introducing self-regularization and motion-centric latent alignment, which facilitate learning an expressive motion latent space and enable effective mapping from text to motion latents (\cref{sec:obj}). Lastly, we provide an in-depth explanation of Reconstructive Error Guidance and inference sampling (\cref{sec:reg}). \cref{fig:framework} provides an overview.

\input{Fig/Fig_01_framework}

\subsection{\modelname Architecture} 
\label{sec:arch}
Given a motion sequence $\mathbf{x}_0 \in \mathbb{R}^{T \times d}$ or a text description $\mathbf{t}$, \modelname either reconstructs the input motion or generates a motion sequence to match a given description. This is achieved through two branches: motion reconstruction and text-to-motion generation, both of which share the Motion Diffusion Model (MDM) \cite{tevet2022mdm} as decoder.

\paragraph{Motion Diffusion Model.}MDM is modeled as a Markov noising chain $\{\mathbf{x}_t\}_{t=0}^T$ with $\mathbf{x}_0$ drawn from the data distribution. The forward diffusion process incrementally adds Gaussian noise: $q(\mathbf{x}_t | \mathbf{x}_{t-1}) = \mathcal{N}\bigl(\mathbf{x}_t;\sqrt{1-\beta_t}\,\mathbf{x}_{t-1},\, \beta_t I\bigr)$. In the reverse process, a denoiser $D$ learns to recover clean motion from a noisy input $\mathbf{x}_t$: $\hat{\mathbf{x}}_t = D(\mathbf{x}_t, t, c)$, where $\hat{\mathbf{x}}_t$ is the motion estimate at timestep $t$ and $c$ denotes conditions.

\paragraph{Motion Reconstruction.}The motion reconstruction branch encodes $\mathbf{x}$ into a latent motion $\mathbf{z}_m$ using a transformer-based motion encoder $E_m(\cdot)$, which takes a special token $\mathbf{s}_m$ and the motion sequence as input. The output $\mathbf{z}_m$ represents the global concept of the sequence. The diffusion decoder $D$ takes $\mathbf{z}_m$, timestep $t$, and noisy motion $\mathbf{x}_t$ to predict clean motion. This process can be expressed as follows:  
\begin{equation} 
\mathbf{z}_m = E_m(\mathbf{s}_m, \mathbf{x}_0), \ \ \ \ \hat{\mathbf{x}}_t = D(\mathbf{x}_t, t, \mathbf{z}_m). 
\end{equation}
Here $t$ denotes the diffusion timestep, which is sampled uniformly as $t \sim \mathcal{U}\{0, \dots, \\ T - 1\}$, where $T$ is the total number of diffusion steps.

\paragraph{Text-to-Motion Generation.}The text-to-motion branch mirrors the motion reconstruction branch, encoding the text embedding $\mathbf{f}_\mathbf{t}$ with a text encoder $E_\mathbf{t}$ and then decoding with $D$: 
\begin{align} 
\mathbf{z}_\mathbf{t} = E_\mathbf{t}(\mathbf{s}_\mathbf{t}, \mathbf{f}_\mathbf{t}), \ \ \ \ \hat{\mathbf{x}}_t = D(\mathbf{x}_t, t, \mathbf{z}_\mathbf{t}).
\end{align} 
Here, $\mathbf{f}_\mathbf{t} \in \mathbb{R}^{L \times d_f}$ is the text embedding at the token-level extracted from $\mathbf{t}$ (where $L$ is the sequence length), $\mathbf{s}_t$ is the special input token, and $\mathbf{z}_\mathbf{t}$ is the latent produced by the text encoder $E_\mathbf{t}$.

\subsection{Optimization Objectives}
\label{sec:obj}

The training objectives of \modelname consist of four key components: reconstruction, text-driven generation, self-regularization, and motion-centric latent alignment. For clarity, we divide these objectives into two categories: (1) reconstruction and text-driven generation, which follow established two-stream approaches \cite{ahuja2019language2pose, petrovich2022temos}, and (2) self-regularization and motion-centric latent alignment, which are our novel contributions aimed at learning a compact yet expressive motion latent space and enabling effective mapping from text to motion latents. In the following, we provide a detailed introduction to the formulation and specific function of each objective.

\paragraph{Reconstruction.} Given the latent motion $\mathbf{z}_m$, the timestep $t$, and the noisy motion $\mathbf{x}_t$, this objective encourages the diffusion model to accurately reconstruct the input motion sequence.
\begin{align}
    L_{\text{rec}} = \mathbb{E}_{\mathbf{x}_0, t}\big[ \lVert D(\mathbf{x}_t, t, \mathbf{z}_m) - \mathbf{x}_0 \rVert_2^2 \big]  = \mathbb{E}_{\mathbf{x}_0, t}\big[ \lVert D(\mathbf{x}_t, t, E_m(\mathbf{s}_m, \mathbf{x}_0)) - \mathbf{x}_0 \rVert_2^2 \big].
\end{align}
This loss jointly trains both the diffusion model and the motion encoder, aiming for a strong motion decoder, an encoder that extracts abstract motion representations, and a compact latent space with essential motion dynamics. 

\paragraph{Text-Driven Generation.}In this objective, the diffusion model learns to generate motion conditioned on the text latent $\mathbf{z}_\mathbf{t}$, timestep $t$, and noisy motion $\mathbf{x}_t$:
\begin{align}
L_{\text{gen}} = \mathbb{E}_{\mathbf{x}_0, t, \mathbf{t}}\big[ \lVert D(\mathbf{x}_t, t, \mathbf{z}_\mathbf{t}) - \mathbf{x}_0 \rVert_2^2 \big]  = \mathbb{E}_{\mathbf{x}_0, t}\big[ \lVert D(\mathbf{x}_t, t, E_\mathbf{t}(\mathbf{s}_\mathbf{t}, \mathbf{f}_\mathbf{t})) - \mathbf{x}_0 \rVert_2^2 \big].
\end{align}
This objective encourages the diffusion model to adapt to conditioning on the text latent manifold, since there are inherent differences between the text and motion manifolds.

\paragraph{Self-Regularization.}This objective can be viewed as a cross-entropy loss operating on the motion latent space. For a batch of size $B$, let the normalized motion latents be $\tilde{\mathbf{z}}_m^i$, and define the similarity $\mathrm{sim}(\tilde{\mathbf{z}}_m^i,\tilde{\mathbf{z}}_m^j) = (\tilde{\mathbf{z}}_m^i)^\top \tilde{\mathbf{z}}_m^j$, which corresponds to the cosine similarity after normalization. With a temperature parameter $\tau=1$, and treating only identical indices as positive pairs, the loss is defined as:
\begin{equation}
L_{\text{sr}} = \frac{1}{B} \sum_{i=1}^{B} 
- \log \frac{\exp(\text{sim}(\tilde{\mathbf{z}}_m^i,\tilde{\mathbf{z}}_m^i)/\tau)}
{\sum_{j=1}^{B} \exp(\text{sim}(\tilde{\mathbf{z}}_m^i,\tilde{\mathbf{z}}_m^j)/\tau)}.
\end{equation}
This loss encourages greater separability among motion latents, producing a broader, more expressive manifold with improved semantic resolution. Consequently, the refined latent space enables more precise mapping from text representations to motion latents in the subsequent alignment objective.

\paragraph{Motion-Centric Latent Alignment.}This objective aligns the text manifold with the motion manifold. Given a paired text description and motion sequence, this objective minimizes the distance between the corresponding text latent $\mathbf{z}_\mathbf{t}$ and motion latent $\mathbf{z}_m$:
\begin{equation}
L_{\text{latent}} = \mathbb{E}_{\mathbf{z}_m, \mathbf{z}_\mathbf{t}} \big[ \lVert \mathbf{z}_\mathbf{t} - (1-\beta)\operatorname{sg}(\mathbf{z}_m) - \beta \mathbf{z}_m \rVert_2^2  \big],
\end{equation}
where $\operatorname{sg}(\cdot)$ is the stop-gradient operator and $\beta$ modulates the flow of gradients to the motion encoder $E_m$. We set $\beta=0.01$ so that \modelname's latent space remains motion-centric while adapting minimally to the text space. This is based on two insights: (1) prioritizing motion space leads to stronger performance than enforcing a fully joint language-motion space, as mapping motion to text sacrifices important motion dynamics, and (2) with end-to-end training, motion latents evolve during alignment. Completely blocking gradients to $E_m$ ($\beta=0$) makes the alignment harder. Thus, a small $\beta$ supports convergence while retaining motion information.

\paragraph{Overall Objective.}The final training objective is a weighted sum:
\begin{equation}
L_{\text{overall}} = L_{\text{rec}} 
+ L_{\text{gen}} 
+ w_{\text{sr}} L_{\text{sr}}
+ w_{\text{latent}} L_{\text{latent}},
\end{equation}
where $w_{\text{sr}}$ and $w_{\text{latent}}$ are hyperparameters that determine the significance of the terms $L_{\text{sr}}$ and $L_{\text{latent}}$, respectively. 

\subsection{Inference}
\label{sec:reg}

\paragraph{Reconstructive Error Guidance.}During training, diffusion models operate exclusively on the canonical data manifold, where the noised input follows $x_t = \sqrt{\bar{\alpha_t}} x_0 + \sqrt{1 - \bar{\alpha_t}} \epsilon$. However, during inference, their predictions often exhibit error patterns and drift away from this manifold. Denoising based on such off-manifold predictions further exacerbates the deviation. In general, diffusion models can cause the sampling path to deviate from the data manifold, leading to degraded sampling quality \cite{li2023error}.

We hypothesize that diffusion models possess an inherent capacity to self-correct such error patterns—a capability analogous to their fundamental ability to recover clean data from noise. However, this corrective potential requires explicit activation and guidance. To harness this intrinsic error-correction capability, we propose an intuitive approach that operates at each denoising step.

Specifically, at inference step $t$, we first reconstruct the prediction from the previous step $t+1$ to capture embedded error patterns explicitly. We then amplify the improvement from the current step's prediction using a residual amplification mechanism. Let $\hat{\mathbf{x}}_{t+1,s}$ denote the final output at step $t+1$. Our method can be formulated as:
\begin{equation}
\hat{\mathbf{x}}_{t,s} = D(\mathbf{x}_t, t, \mathbf{z}_\mathbf{t}) + w \left( D(\mathbf{x}_t, t, \mathbf{z}_\mathbf{t}) - D(\mathbf{x}_t, t, \mathbf{z}_{m,t+1}) \right),
\end{equation}
where $\mathbf{z}_{m,t+1} = E_m(\mathbf{s}_m, \hat{\mathbf{x}}_{t+1,s})$ represents the reconstructed motion latent from the previous step, and $w \geq 0$ is a weighting coefficient that controls the amplification strength of the residual correction term. We term this inference strategy Reconstructive Error Guidance (REG).

\paragraph{Inference Sampling.}Finally, during inference, we combine REG with the commonly used classifier-free guidance (CFG) for sampling. The final output for each denoising step $t$, denoted as $\hat{\mathbf{x}}_{t,s}$, is computed as:
\begin{align}
\hat{\mathbf{x}}_{t,s}
 = & D(\mathbf{x}_t, t, \mathbf{z}_\mathbf{t}) 
 + w_1 \underbrace{\left( D(\mathbf{x}_t, t, \mathbf{z}_\mathbf{t})
           - D(\mathbf{x}_t, t, \mathbf{z}_{m,t+1}) \right)}_{\text{REG term}} 
           \nonumber \\
& + w_2 \underbrace{\left( D(\mathbf{x}_t, t, \mathbf{z}_\mathbf{t})
           - D(\mathbf{x}_t, t, \varnothing) \right)}_{\text{CFG term}},
\label{eq:reg_cfg}
\end{align}
where the final term represents the standard CFG residual between conditional and unconditional predictions (with unconditional input denoted by $\varnothing$). Here, $w_1$ and $w_2$ respectively control the influence of REG and CFG.

%% file: Fig/Fig_01_framework.tex
\newcommand{\frameworkCaption}{
\textbf{Overview of \modelname.} 
During training, \modelname learns a motion latent space through motion reconstruction, with self-regularization to encourage better separability between motion latents, resulting in improved semantic resolution. The text latents from the text encoder are drawn closer to corresponding motion latents through motion-centric latent alignment. At each inference step, given the last step prediction $\hat{\mathbf{x}}_{t+1,s}$ and text description, \modelname first encodes them into latents $\mathbf{z}_{m,t+1}$ and $\mathbf{z}_\mathbf{t}$. Then, these latents, together with a zero vector $\varnothing$ and input noise, are fed into the diffusion model to separately obtain reconstruction, text-driven prediction, and unconditional prediction. These outputs are combined to produce the final output. Best viewed in color.
}

\begin{figure}
    \centering
    \includegraphics[width=1.0\linewidth]{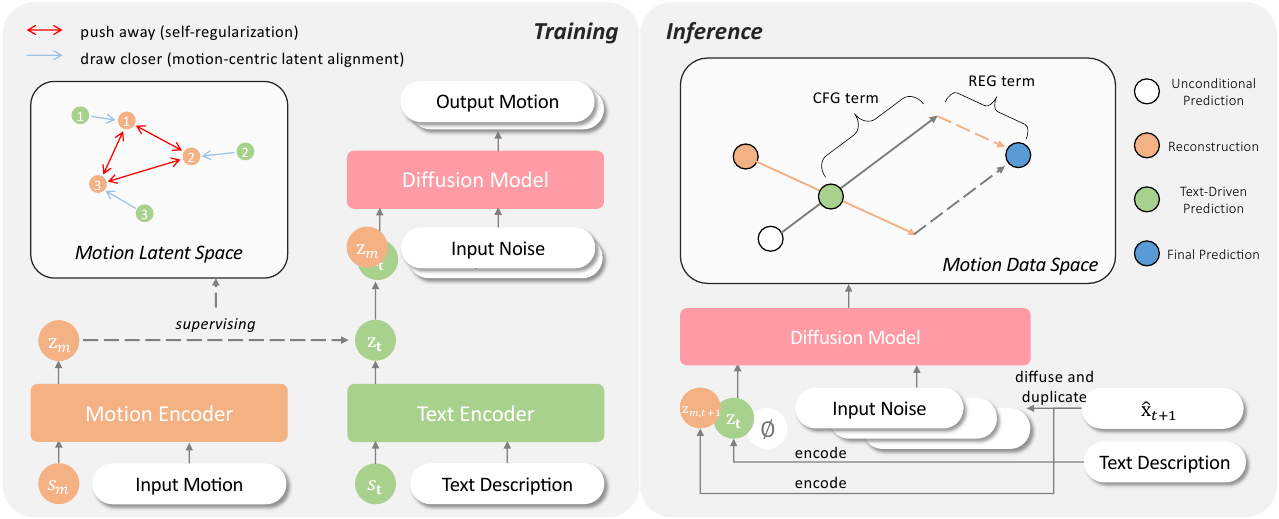}
    \caption{\frameworkCaption}
    \label{fig:framework}
\end{figure}

%% file: Sec/Sec_04_experiments.tex
\section{Experiment}

\subsection{Datasets and Metrics}

\paragraph{Datasets.}HumanML3D \cite{guo2022humanml3d} is a large-scale text–motion dataset containing 14,616 motion sequences from AMASS \cite{amass}, each annotated with 44,970 sequence-level textual descriptions. By comparison, the KIT-ML dataset \cite{plappert2016kit} is smaller, offering 3,911 motion sequences and 6,353 textual descriptions. For both datasets, we use the standard redundant motion representation \cite{guo2022humanml3d}, which includes joint velocities, positions, and rotations.

\paragraph{Metrics.}We assess the generated motions with five complementary metrics. R-Precision and Multimodal-Dist measure the semantic alignment between generated motions and text descriptions. Fréchet Inception Distance (FID) evaluates the distributional similarity between generated motions and the ground truth in a learned latent space. Diversity (Div) quantifies the variability within the generated motion set, while MultiModality (MM) captures the average variance among motions conditioned on the same description.

\input{Tab/Tab_01_sota_ml3d}

\input{Tab/Tab_02_sota_kit}

\subsection{Implementation Details}
\label{sec4.2}

We adopt exactly the same text and motion encoders as those in TEMOS \cite{petrovich2022temos}. Both of them are implemented as 6-layer, encoder-only transformers. The text encoder takes the text embeddings extracted by DistilBERT \cite{sanh2019distilbert} as input. 
The latent dimensionality is set to 256 for HumanML3D and 192 for KIT-ML. For the diffusion model that generates motion sequences from the latents, we use the MDM architecture \cite{tevet2022mdm}, consisting of an 8-layer, encoder-only transformer backbone with a latent size of 512. Training is performed with a batch size of 64, a learning rate of 0.0001, and the AdamW optimizer. Models are trained for 450K steps on HumanML3D and 400K steps on KIT-ML. 
We set the loss weights $w_{\text{sr}}$ to 1 and $w_{\text{latent}}$ to 0.5.
During training, we set the diffusion timestep $T$ to 50, with 10\% of conditional latents replaced by zero vectors for classifier-free guidance. For inference on HumanML3D, we use 20 denoising steps, spaced linearly in $[0, \ldots, T-1]$, resulting in 20 inference steps. The weights for REG and classifier-free guidance are set to 5.0 and 1.5, respectively.

\subsection{Comparison with State-of-the-Art Methods}

We quantitatively compare \modelname with state-of-the-art (SOTA) methods on HumanML3D and KIT-ML. The results are shown in \cref{tab:ml3d} and \cref{tab:kit}, respectively. As demonstrated in \cref{tab:ml3d}, \modelname achieves SOTA performance on the most widely used HumanML3D benchmark. Compared with diffusion-based methods, \modelname shows a substantial improvement in FID and achieves near-SOTA performance in semantic accuracy as measured by R-Precision, ranking just behind Salad. When compared to VQ-VAE-based approaches, \modelname surpasses them in semantic accuracy and achieves highly competitive FID—\textit{a feat not previously attained by diffusion-based methods}. \modelname thus demonstrates that diffusion-based motion generation models can reach state-of-the-art FID levels.

On the KIT-ML dataset, the limited training scale poses significant challenges for generative modeling, particularly for diffusion-based architectures. Consequently, \modelname, similar to the leading diffusion-based method Salad \cite{hong2025salad}, exhibits a performance variance compared to larger benchmarks. Nevertheless, \modelname remains highly competitive, achieving strong performance in both motion realism and semantic alignment.

\paragraph{Inference Efficiency.}\highlight{Although REG adds one reconstruction forward per step, \modelname stays efficient: with full REG at 20 steps it runs at 0.398\,s AITS—faster than MDM at 50 steps (0.490\,s)—yet reaches a far lower FID, and restricting REG to the first 6 steps adds only $26\%$ time for a $71\%$ FID drop ($0.132{\to}0.038$); see \aref{sup:effi} for more details.}

\subsection{Ablation Studies}
\label{sec:4.4abl}

To assess the impact of key design choices within \modelname, we conduct comprehensive ablation studies on HumanML3D. Specifically, these studies include: (1) \textit{Incremental Experiments}—starting from a baseline model, we progressively introduce key design components, culminating in the complete \modelname;
and (2) \textit{Guidance Evaluation}—we examine the effectiveness of our proposed Reconstructive Error Guidance and the additional benefits achieved when it is combined with classifier-free guidance (CFG). In addition, we provide further experiments on the influence of internal loss parameters $\beta$ and $\tau$, loss weights $w_\text{st}$ and $w_\text{latent}$, and encoder latent dimensionality in \aref{sup:hyper}.

\input{Tab/Tab_03_ablation}

\paragraph{Incremental Experiments.}\cref{tab:ablation} presents the results of our incremental
studies. To rigorously assess the contribution of each component, we begin with a clean baseline consisting of \modelname's text encoder and diffusion model only. We keep these modules exactly the same as those in \modelname and progressively add key components. The baseline demonstrates limited performance, partly due to the challenging inference setting of only 20 steps. Introducing the motion encoder $E_m$—which forms a dual-branch architecture, similar to a direct application of Language2pose \cite{ahuja2019language2pose} to diffusion models—provides a modest improvement, suggesting that implicitly learning a joint language-motion space is suboptimal. Incorporating $L_\text{latent}$ delivers substantial further gains, though still falls short of state-of-the-art performance. Adding $L_\text{sr}$ leads to results that surpass most diffusion-based approaches reported in \cref{tab:ml3d}. Finally, enabling REG elevates \modelname to state-of-the-art performance. Collectively, these findings demonstrate the necessity and effectiveness of each design choice. 

\input{Tab/Tab_05_infer}

\paragraph{Guidance Evaluation.}\cref{tab:infer} presents the results of our guidance evaluation. The findings show that classifier-free guidance (CFG) substantially improves semantic accuracy, as measured by R-Precision. In contrast, our proposed reconstructive error guidance (REG) notably enhances the overall realism of the generated motion, reflected by better FID scores. Furthermore, combining both strategies enables \modelname to achieve state-of-the-art performance.

\subsection{Comprehensive Comparison of Latent Space Training Strategies}

In \cref{sec:4.4abl}, we conducted a preliminary comparison between directly applying dual-branch architecture to diffusion models and our reconstruction-anchored approach to demonstrate the superiority of \modelname over traditional two-stream methods. Our reconstruction-anchored approach is characterized by two key components: (1) self-regularization performed in the motion latent space, and (2) motion-centric latent alignment that aligns text representations with the motion latent space, where this motion latent space is learned through motion reconstruction. In this section, we conduct a more comprehensive comparison across multiple variants to provide stronger empirical evidence supporting our approach.

First, we introduce several variants for comparison:
\begin{itemize}
    \item {Two-stream baseline (Variant A)}: This variant represents the most basic approach where reconstruction and generation branches share a latent space without directly imposing any loss on the latent space, only implicitly learning a joint language–motion embedding space. A representative method is Language2Pose~\cite{ahuja2019language2pose}.
    
    \item {Bidirectional latent alignment (Variant B and C)}: This variant directly applies alignment loss between motion and text latent variables in the latent space. A representative method is TEMOS~\cite{petrovich2022temos}. We provide two weight configurations: a loss weight of 1.0 ({Variant B}) and a loss weight of $10^{-5}$ ({Variant C}), where the latter reproduces the parameter settings from the original TEMOS paper.
    
    \item {Cross-modal contrastive learning (Variant D)}: This variant applies cross-modal contrastive learning in the latent space, which constitutes one key objective of TMR \cite{petrovich2023tmr} and KinMo \cite{zhang2025kinmo}.
    
    \item {Cross-modal contrastive learning \& bidirectional latent alignment (Variant E)}: This variant simultaneously applies both cross-modal contrastive learning and bidirectional latent alignment in the latent space. TMR is a representative method using this approach. To reproduce TMR's parameter settings, we set the contrastive learning temperature $\tau=0.1$ with a weight of 0.1, and the bidirectional latent alignment weight to $10^{-5}$.
    
    \item {Ours}: This variant employs self-regularization and motion-centric alignment.
\end{itemize}
Implementation details of these variants, including network architectures, hyperparameters, and inference configurations, are provided in \aref{sup:latent_comparison_details}.

\input{Tab/Tab_09_embeddingspace}

As shown in \cref{tab:embeddingspace}, directly applying bidirectional latent alignment (Variant B) and cross-modal contrastive learning (Variant D) to the diffusion-based two-stream baseline fails to yield performance improvements and actually results in degradation. This indicates that incorporating these techniques into two-stream methods requires careful hyperparameter tuning. When employing hyperparameters empirically determined from prior work (Variant C \cite{petrovich2022temos} and Variant E \cite{petrovich2023tmr}), performance gains are observed. Notably, Variant E achieves R-Precision comparable to our approach. However, a substantial gap remains between Variant E and our method in terms of FID. This suggests that our motion-centric approach better preserves the intrinsic structure of the motion space, demonstrating superior motion fidelity.

\input{Fig/Fig_02_qualitative}

\subsection{Qualitative Evaluation}
\label{sec:qualitative}

We compare the qualitative results of \modelname with those generated by three representative or leading text-to-motion methods: MDM \cite{tevet2022mdm}, MoMask \cite{guo2024momask}, and Salad \cite{hong2025salad}. \Cref{fig:qualitative} illustrates four groups of comparisons, with the corresponding text descriptions shown below each group. The text prompts are sampled from the test set. Since each prompt consists of multiple actions, this setting poses a significant challenge for accurate motion generation. The results show that our method achieves a high degree of semantic accuracy and realism. Additional qualitative examples are provided in the \textbf{supplementary video}, and we further validate the effectiveness of our method from a perceptual perspective through a \textbf{user study} in \aref{sup:user}.

%% file: Tab/Tab_01_sota_ml3d.tex
\begin{table*}[ht]
\centering
\newcolumntype{M}[1]{>{\centering\arraybackslash}m{#1}}

    \caption{Quantitative results of text-to-motion generation on the HumanML3D test set. Methods are grouped into VQ-VAE-based and diffusion-based categories. In each group, the best results are highlighted in \textbf{bold}, while the second-best results are \underline{underlined}. Notably, \modelname achieves an FID of 0.032, surpassing prior diffusion-based methods and outperforming most VQ-VAE-based models.}
  \begin{adjustbox}{center, width=0.8\textwidth}
  \begin{tabular}{cl|c|ccc|ccc}
\toprule
\multirow{2}{*}{} &\multirow{2}{*}{Method} & \multirow{2}{*}{FID$\downarrow$} & \multicolumn{3}{c|}{R-Precision$\uparrow$} & \multirow{2}{*}{MM Dist$\downarrow$} & \multirow{2}{*}{Diversity} & \multirow{2}{*}{MM} \\

& & & {Top 1} & {Top 2} & {Top 3}  \\
\midrule

& Ground Truth   & $0.002^{\pm .000}$ & $0.511^{\pm .003}$ & $0.703^{\pm .003}$ & $0.797^{\pm .002}$ & $2.974^{\pm .008}$ & $9.503^{\pm .065}$ & - \\ \hline
\multirow{8}{*}{\rotatebox[origin=c]{90}{VQ-VAE-based}} & T2M-GPT \cite{zhang2023t2mgpt}   & $0.116^{\pm .004}$ & $0.491^{\pm .003}$ & $0.680^{\pm .003}$ & $0.775^{\pm .002}$ & $3.118^{\pm .011}$ & $9.761^{\pm .081}$ & $1.856^{\pm .011}$ \\
&MMM \cite{pinyoanuntapong2024mmm} & $0.080{ }^{ \pm .003}$ & $0.504^{ \pm .003}$ & $0.696^{ \pm .003}$ & $0.794^{ \pm .002}$ & $2.998^{ \pm .007}$ & $9.411^{ \pm .058}$ & $1.164^{ \pm .041}$ \\
&MoMask \cite{guo2024momask}     & ${0.045}^{\pm .002}$ & $0.521^{\pm .002}$ & $0.713^{\pm .002}$ & $0.807^{\pm .002}$ & $2.958^{\pm .008}$ & -                 & $1.241^{\pm .040}$ \\
&BAMM \cite{pinyoanuntapong2024bamm} & $0.055^{ \pm .002}$ & $0.525^{ \pm .002}$ & $0.720^{ \pm .003}$ & $0.814^{ \pm .003}$ & $2.919{ }^{ \pm .008}$ & $9.717^{ \pm .089}$ & $1.687^{ \pm .051}$ \\
&MoGenTS \cite{yuan2024mogents} & $\underline{0.033}^{ \pm .001}$ & $0.529^{ \pm .003}$ & $0.719^{ \pm .002}$ & $0.812^{ \pm .002}$ & $\underline{2.867}^{ \pm .006}$ & $9.570^{ \pm .077}$ & - \\
&BAD \cite{hosseyni2025bad} & $0.065^{ \pm .003}$ & $0.517{ }^{ \pm .002}$ & $0.713{ }^{ \pm .003}$ & $0.808^{ \pm .003}$ & $2.901^{ \pm .008}$ & $9.694^{ \pm .068}$ & $1.194^{ \pm .044}$ \\
&KinMo \cite{zhang2025kinmo} & $0.039^{ \pm .003}$ & $\underline{0.532}^{ \pm .002}$ & $\underline{0.724}^{ \pm .003}$ & $\underline{0.821}^{ \pm .003}$ & $2.901^{ \pm .010}$ & $9.674^{ \pm .058}$ & $1.321^{ \pm .039}$ \\
&LaMP \cite{li2025lamp} & $\mathbf{0.032^{ \pm .002}}$ & $\mathbf{0.557}^{ \pm .003}$ & $\mathbf{0.751}^{ \pm .002}$ & $\mathbf{0.843}^{ \pm .001}$ & $\mathbf{2.759}^{ \pm .007}$ & $9.571^{ \pm .069}$ & - \\
\hline
\multirow{13}{*}{\rotatebox[origin=c]{90}{Diffusion-based}} &MDM \cite{tevet2022mdm}       & $0.489^{\pm .025}$ & $0.418^{\pm .005}$ & $0.604^{\pm .001}$ & $0.707^{\pm .004}$ & $3.360^{\pm .023}$ & $9.450^{\pm .066}$ & $2.860^{\pm 1.11}$ \\
&\highlight{MotionDiffuse \cite{zhang2022motiondiffuse}} & \highlight{$0.630^{\pm .001}$} & \highlight{$0.491^{\pm .001}$} & \highlight{$0.681^{\pm .001}$} & \highlight{$0.782^{\pm .001}$} & \highlight{$3.113^{\pm .001}$} & \highlight{$9.410^{\pm .049}$} & \highlight{$1.553^{\pm .040}$} \\
&MLD \cite{chen2023executing}       & $0.473^{\pm .013}$ & $0.481^{\pm .003}$ & $0.673^{\pm .003}$ & $0.772^{\pm .002}$ & $3.196^{\pm .010}$ & $9.724^{\pm .082}$ & $2.413^{\pm .079}$ \\
&ReMoDiffuse \cite{zhang2023remodiffuse}       & $0.103^{\pm .004}$ & $0.510^{\pm .005}$ & $0.698^{\pm .006}$ & $0.795^{\pm .004}$ & $2.974^{\pm .016}$ & $9.018^{\pm .075}$ & $1.795^{\pm .043}$ \\
&FineMoGen \cite{zhang2023finemogen} & $0.151^{\pm .008}$ & $0.504^{\pm .002}$ & $0.690^{\pm .002}$ & $0.784^{\pm .002}$ & $2.998^{\pm .008}$ & $9.263^{\pm .094}$ & $2.696^{\pm .079}$ \\
&MotionLCM \cite{dai2024motionlcm}     & $0.304^{\pm .012}$ & $0.502^{\pm .003}$ & $0.698^{\pm .002}$ & $0.798^{\pm .002}$ & $3.012^{\pm .007}$ & $9.607^{\pm .066}$                 & $2.259^{\pm .092}$ \\
&\highlight{MotionLCM v2 \cite{motionlcm-v2}} & \highlight{$0.056^{\pm .003}$} & \highlight{$0.553^{\pm .003}$} & \highlight{$0.746^{\pm .002}$} & \highlight{$0.837^{\pm .002}$} & \highlight{$2.773^{\pm .009}$} & \highlight{$9.598^{\pm .067}$} & \highlight{$1.758^{\pm .056}$} \\
&StableMoFusion \cite{stablemofusion}  & $0.098^{ \pm .003}$ & $0.553^{ \pm .003}$ & $0.748^{ \pm .002}$ & $0.841^{ \pm .002}$ & - & $9.748^{ \pm .092}$ & $1.774^{ \pm .051}$ \\
&CLoSD \cite{tevet2025closd}            & $0.283^{\pm .000}$ & $0.464^{\pm .000}$ & $0.668^{\pm .000}$ & $0.777^{\pm .000}$ & $3.150^{\pm .000}$ & $9.210^{\pm .000}$                 & - \\
&Salad \cite{hong2025salad}            & ${0.076}^{\pm .002}$ & $\mathbf{0.581}^{\pm .003}$ & $\mathbf{0.769}^{\pm .003}$ & $\mathbf{0.857}^{\pm .002}$ & $\mathbf{2.649}^{\pm .009}$ & $9.696^{\pm .096}$                 & $1.751^{\pm .062}$ \\
&sMDM \cite{bae2025less}            & $0.130^{\pm .000}$ & $0.494^{\pm .000}$ & $0.682^{\pm .000}$ & $0.776^{\pm .000}$ & $3.051^{\pm .000}$ & $9.663^{\pm  .000}$                 & - \\
&\highlight{COME \cite{lyucome}} & \highlight{$\underline{0.041}^{\pm .002}$} & \highlight{$0.526^{\pm .003}$} & \highlight{$0.723^{\pm .002}$} & \highlight{$0.816^{\pm .002}$} & \highlight{$2.898^{\pm .006}$} & \highlight{$9.532^{\pm .062}$} & \highlight{$1.704^{\pm .059}$} \\
&\textbf{\modelname} (Ours) & $\mathbf{0.032^{\pm .002}}$ & $\underline{0.561}^{\pm .003}$ & $\underline{0.751}^{\pm .002}$ & $\underline{0.839}^{\pm .002}$ & $\underline{2.716}^{\pm .007}$ & $9.487^{\pm .084}$   & $1.142^{\pm .038}$ \\
\bottomrule
\end{tabular}
\end{adjustbox}

  \label{tab:ml3d}
\end{table*}

%% file: Tab/Tab_02_sota_kit.tex
\begin{table*}[ht]
\centering
  \caption{Quantitative results of text-to-motion generation on the KIT-ML test set. Methods are grouped into VQ-VAE-based and diffusion-based categories. In each group, the best results are highlighted in \textbf{bold}, while the second-best results are \underline{underlined}. Notably, on this challenging, smaller dataset, \modelname achieves the most balanced performance among diffusion models, ranking second in both FID and R-Precision. It substantially outperforms the FID leader (ReMoDiffuse) in R-Precision and the R-Precision leader (Salad) in FID.}
  \begin{adjustbox}{center, width=0.8\textwidth}
  \begin{tabular}{cl|c|ccc|ccc}
\toprule
\multirow{2}{*}{} &\multirow{2}{*}{Method} & \multirow{2}{*}{FID$\downarrow$} & \multicolumn{3}{c|}{R-Precision$\uparrow$} & \multirow{2}{*}{MM Dist$\downarrow$} & \multirow{2}{*}{Diversity} & \multirow{2}{*}{MM} \\
& & & {Top 1} & {Top 2} & {Top 3}  \\
\midrule

&Ground Truth    & $0.031^{\pm .004}$ & $0.424^{\pm .005}$ & $0.649^{\pm .006}$ & $0.779^{\pm .006}$ & $2.788^{\pm .012}$ & $11.08^{\pm .097}$ & - \\ \hline

\multirow{7}{*}{\rotatebox[origin=c]{90}{VQ-VAE-based}} 
&T2M-GPT \cite{zhang2023t2mgpt}   & $0.512^{\pm .029}$ & $0.416^{\pm .006}$ & $0.627^{\pm .006}$ & $0.745^{\pm .006}$ & $3.007^{\pm .023}$ & $10.92^{\pm .108}$ & $1.856^{\pm .011}$ \\
&MMM \cite{pinyoanuntapong2024mmm} & $0.316^{ \pm .028}$ & $0.404^{ \pm .005}$ & $0.621^{ \pm .005}$ & $0.744^{ \pm .004}$ & $2.977^{ \pm .019}$ & $10.91^{ \pm .101}$ & $1.232^{ \pm .039}$ \\
&MoMask \cite{guo2024momask}     & $0.204^{\pm .011}$ & $0.433^{\pm .007}$ & $0.656^{\pm .005}$ & $0.781^{\pm .005}$ & $2.779^{\pm .022}$ & -                 & $1.131^{\pm .043}$ \\
&BAMM \cite{pinyoanuntapong2024bamm} & $0.183^{ \pm .013}$ & $0.438^{ \pm .009}$ & $0.661^{ \pm .009}$ & $0.788^{ \pm .005}$ & ${2.723}^{\pm .026}$ & $11.01^{ \pm .094}$ & $1.609^{ \pm .065}$ \\
&MoGenTS \cite{yuan2024mogents} & $\underline{0.143}^{ \pm .004}$ & $\underline{0.445}^{ \pm .006}$ & $\underline{0.671}^{ \pm .006}$ & $\underline{0.797}^{ \pm .005}$ & $\underline{2.711}^{ \pm .024}$ & $10.92^{ \pm .090}$ & - \\
&BAD \cite{hosseyni2025bad} & $0.221^{ \pm .012}$ & $0.417^{ \pm .006}$ & $0.631^{ \pm .006}$ & $0.750^{ \pm .006}$ & $2.941^{ \pm .025}$ & $11.00^{ \pm .100}$ & $1.170^{ \pm .047}$ \\
&LaMP \cite{li2025lamp} & $\mathbf{0.141}^{ \pm .013}$ & $\mathbf{0.479}^{ \pm .006}$ & $\mathbf{0.691}^{ \pm .005}$ & $\mathbf{0.826}^{ \pm .005}$ & $\mathbf{2.704}^{ \pm .018}$ & $10.93^{ \pm .101}$ & - \\

\hline

\multirow{7}{*}{\rotatebox[origin=c]{90}{Diffusion-based}}
& MDM \cite{tevet2022mdm}       & $0.547^{\pm .070}$ & $0.404^{\pm .002}$ & $0.616^{\pm .013}$ & $0.737^{\pm .005}$ & $3.074^{\pm .018}$ & $10.75^{\pm .203}$ & $1.806^{\pm .180}$ \\
&MLD \cite{chen2023executing}       & $0.404^{\pm .027}$ & $0.390^{\pm .008}$ & $0.609^{\pm .008}$ & $0.734^{\pm .007}$ & $3.204^{\pm .027}$ & $10.80^{\pm .117}$ & $2.192^{\pm .071}$ \\
&ReMoDiffuse \cite{zhang2023remodiffuse}       & $\mathbf{0.155}^{\pm .006}$ & $0.427^{\pm .014}$ & $0.641^{\pm .004}$ & $0.765^{\pm .055}$ & $2.814^{\pm .012}$ & $10.80^{\pm .105}$ & $1.239^{\pm .028}$ \\
&FineMoGen \cite{zhang2023finemogen} & $0.178^{\pm .007}$ & $0.432^{\pm .006}$ & $0.649^{\pm .005}$ & $0.772^{\pm .006}$ & $2.869^{\pm .014}$ & $10.85^{\pm .115}$ & $1.877^{\pm .093}$ \\
&StableMoFusion \cite{stablemofusion} & $0.258^{ \pm .029}$ & $0.445^{ \pm .006}$ & $0.660^{ \pm .005}$ & $0.782^{ \pm .004}$ & - & $10.94^{ \pm .077}$ & $1.362^{ \pm .062}$ \\
&Salad \cite{hong2025salad}            & $0.296^{\pm .012}$ & $\mathbf{0.477}^{\pm .006}$ & $\mathbf{0.711}^{\pm .005}$ & $\mathbf{0.828}^{\pm .005}$ & $\mathbf{2.585}^{\pm .016}$ & $11.10^{\pm .095}$                 & $1.004^{\pm .040}$ \\
&\textbf{\modelname} (Ours) & $\underline{0.172}^{\pm .010}$ & $\underline{0.464}^{\pm .006}$ & $\underline{0.684}^{\pm .006}$ & $\underline{0.803}^{\pm .005}$ & $\underline{2.653}^{\pm .024}$ & $11.15^{\pm .101}$         & $1.010^{\pm .043}$ \\

\bottomrule
\end{tabular}
\end{adjustbox}

\label{tab:kit}
\end{table*}

%% file: Tab/Tab_03_ablation.tex
\begin{table*}[t]
\centering
\scriptsize
\caption{Incremental ablation experiments on the key components of \modelname. We evaluate the impact of the motion reconstruction branch, Self-Regularization ($L_{sr}$), Motion-Centric Latent Alignment ($L_{latent}$), and Reconstructive Error Guidance (REG). The results show that each component makes an essential contribution, with the full model achieving the best performance on both FID and R-Precision. \highlight{In each column, the best result is highlighted in \textbf{bold} and the second-best is \underline{underlined}.}}
\begin{tabular*}{\linewidth}{@{\extracolsep{\fill}}cccc|c|ccc|cc}
\toprule
\multicolumn{4}{c|}{Components} & \multirow{2}{*}{FID$\downarrow$} & \multicolumn{3}{c|}{R-Precision$\uparrow$} & \multirow{2}{*}{MM D$\downarrow$} & \multirow{2}{*}{Diversity}\\
$E_m$ & $L_{\text{latent}}$ & $L_{\text{sr}}$ & REG & & Top 1 & Top 2 & Top 3 & & \\
\midrule
 &  &  &  & $0.786^{\pm .016}$ & $0.417^{\pm .002}$ & $0.613^{\pm .002}$ & $0.729^{\pm .003}$ & $3.433^{\pm .012}$ & $10.063^{\pm .076}$ \\
\checkmark &  &  &  & $0.624^{\pm .013}$ & $0.493^{\pm .004}$ & $0.695^{\pm .002}$ & $0.800^{\pm .002}$ & $3.045^{\pm .013}$ & $10.188^{\pm .096}$ \\
\checkmark & \checkmark &  & & $0.187^{\pm .005}$ & $0.530^{\pm .002}$ & $0.720^{\pm .002}$ & $0.814^{\pm .002}$ & $2.885^{\pm .008}$ & $9.703^{\pm .077}$  \\
\checkmark & \checkmark & \checkmark &  & $\underline{0.132}^{\pm .005}$ & $\mathbf{0.561}^{\pm .002}$ & $\mathbf{0.752}^{\pm .002}$ & $\underline{0.838}^{\pm .002}$ & $\underline{2.744}^{\pm .006}$ & $9.754^{\pm .067}$ \\
\checkmark & \checkmark & \checkmark & \checkmark & ${\mathbf{0.032}^{\pm .002}}$ & ${\mathbf{0.561}}^{\pm .003}$ & ${\underline{0.751}}^{\pm .002}$ & ${\mathbf{0.839}}^{\pm .002}$ & ${\mathbf{2.716}}^{\pm .007}$ & $9.487^{\pm .084}$ \\
\bottomrule
\end{tabular*}

\label{tab:ablation}
\end{table*}

%% file: Tab/Tab_05_infer.tex
\begin{table}
\centering
\setlength{\tabcolsep}{6pt}
\scriptsize
\caption{Effect of Reconstructive Error Guidance (REG) and classifier-free guidance (CFG). $w_1$ and $w_2$ respectively control the influence of REG and CFG. The results demonstrate that CFG substantially improves semantic accuracy (higher R-Precision), while REG notably enhances motion realism (lower FID). Furthermore, combining both strategies enables RAM to achieve state-of-the-art performance. \highlight{In each column, the best result is highlighted in \textbf{bold} and the second-best is \underline{underlined}.}}
\begin{tabular*}{\linewidth}{@{\extracolsep{\fill}} cc|c|ccc|cc}
\toprule
\multirow{2}{*}{$w_1$} & \multirow{2}{*}{$w_2$} & \multirow{2}{*}{FID$\downarrow$} & \multicolumn{3}{c|}{R-Precision$\uparrow$} & \multirow{2}{*}{MM D$\downarrow$} & \multirow{2}{*}{Diversity} \\
 & & & Top 1 & Top 2 & Top 3 & & \\
\midrule
0.0 & 0.0 & $0.297^{\pm .010}$ & $0.529^{\pm .002}$ & $0.719^{\pm .002}$ & $0.812^{\pm .002}$ & $2.912^{\pm .007}$ & $9.707^{\pm .072}$ \\
3.0 & 0.0 & $\underline{0.088}^{\pm .005}$ & $0.542^{\pm .002}$ & $0.732^{\pm .002}$ & $0.825^{\pm .002}$ & $2.808^{\pm .007}$ & $9.418^{\pm .075}$ \\
4.0 & 0.0 & $0.097^{\pm .005}$ & $0.539^{\pm .002}$ & $0.729^{\pm .001}$ & $0.823^{\pm .001}$ & $2.817^{\pm .007}$ & $9.328^{\pm .071}$ \\
5.0 & 0.0 & $0.128^{\pm .006}$ & $0.537^{\pm .002}$ & $0.726^{\pm .001}$ & $0.820^{\pm .001}$ & $2.837^{\pm .008}$ & $9.242^{\pm .068}$ \\
0.0 & 1.5 & $0.132^{\pm .005}$ & $0.561^{\pm .002}$ & $0.752^{\pm .002}$ & $0.838^{\pm .002}$ & $2.744^{\pm .006}$ & $9.754^{\pm .067}$ \\
0.0 & 2.5 & $0.107^{\pm .004}$ & $\mathbf{0.563}^{\pm .002}$ & $\underline{0.753}^{\pm .002}$ & $\mathbf{0.840}^{\pm .002}$ & $2.729^{\pm .006}$ & $9.711^{\pm .066}$ \\
0.0 & 3.5 & $0.097^{\pm .004}$ & $\underline{0.562}^{\pm .002}$ & $\mathbf{0.754}^{\pm .002}$ & $\mathbf{0.840}^{\pm .002}$ & $\underline{2.727}^{\pm .006}$ & $9.668^{\pm .067}$ \\
0.0 & 4.5 & $0.095^{\pm .003}$ & $\underline{0.562}^{\pm .002}$ & $0.752^{\pm .002}$ & $\underline{0.839}^{\pm .002}$ & $2.734^{\pm .007}$ & $9.633^{\pm .068}$ \\
5.0 & 1.5 & ${\mathbf{0.032}^{\pm .002}}$ & ${0.561}^{\pm .003}$ & ${0.751}^{\pm .002}$ & ${\underline{0.839}}^{\pm .002}$ & ${\mathbf{2.716}}^{\pm .007}$ & $9.487^{\pm .084}$   \\
\bottomrule
\end{tabular*}

\label{tab:infer}
\end{table}

%% file: Tab/Tab_09_embeddingspace.tex
\begin{table*}
\centering
\setlength{\tabcolsep}{7pt}
\scriptsize
\caption{Comparison of latent space training strategies. \textbf{Variant A}: Two-stream baseline. \textbf{Variant B}: Bidirectional latent alignment (weight=1.0). \textbf{Variant C}: Bidirectional latent alignment (weight=$10^{-5}$). \textbf{Variant D}: Cross-modal contrastive learning. \textbf{Variant E}: Cross-modal contrastive learning \& bidirectional latent alignment. The results indicate that while Variant E achieves semantic accuracy (R-Precision) comparable to ours, Our method significantly outperforms it in terms of FID. This demonstrates that our motion-centric alignment better preserves the intrinsic structure of the motion manifold compared to bidirectional or contrastive approaches.}
\begin{tabular*}{\linewidth}{@{\extracolsep{\fill}} c|c|ccc|cc}
\toprule
\multirow{2}{*}{Variant} & \multirow{2}{*}{FID$\downarrow$} & \multicolumn{3}{c|}{R-Precision$\uparrow$} & \multirow{2}{*}{MM Dist$\downarrow$} & \multirow{2}{*}{Diversity}\\
 & & Top 1 & Top 2 & Top 3 & & \\
\midrule
Variant A & $0.624^{\pm .013}$ & $0.493^{\pm .004}$ & $0.695^{\pm .002}$ & $0.800^{\pm .002}$ & $3.045^{\pm .013}$ & $10.188^{\pm .096}$ \\
Variant B & $0.776^{\pm .015}$ & $0.415^{\pm .002}$ & $0.610^{\pm .003}$ & $0.724^{\pm .002}$ & $3.430^{\pm .012}$ & $10.029^{\pm .084}$ \\
Variant C & $0.540^{\pm .008}$ & $0.494^{\pm .002}$ & $0.696^{\pm .002}$ & $0.799^{\pm .002}$ & $3.020^{\pm .008}$ & $10.029^{\pm .064}$ \\
Variant D & $0.708^{\pm .015}$ & $0.419^{\pm .004}$ & $0.626^{\pm .003}$ & $0.740^{\pm .003}$ & $3.349^{\pm .011}$ & $10.107^{\pm .077}$ \\
Variant E & $0.218^{\pm .006}$ & $0.558^{\pm .003}$ & $0.750^{\pm .002}$ & $0.839^{\pm .002}$ & $2.767^{\pm .010}$ & $9.820^{\pm .074}$ \\
Ours & $0.132^{\pm .005}$ & $0.561^{\pm .002}$ & $0.752^{\pm .002}$ & $0.838^{\pm .002}$ & $2.744^{\pm .006}$ & $9.754^{\pm .067}$ \\
\bottomrule
\end{tabular*}

\label{tab:embeddingspace}
\end{table*}

%% file: Fig/Fig_02_qualitative.tex
\newcommand{\qualitativeCaption}{
Qualitative evaluation on the HumanML3D Dataset. Actions corresponding to the text are highlighted with \textcolor{ForestGreen}{green} dashed lines, while unnatural artifacts are indicated with \textcolor{red}{red} dashed lines. 
It can be observed that baseline methods often fail to faithfully execute the entire set of actions described in the text. For example, in the second row (``a person walks up and tosses something''), most methods only execute the walking motion. Some outputs also exhibit distortions, such as unnatural drifting (in the third row, MoMask during sitting down and Salad during standing up) and error patterns (in the first row, MoMask’s hands move erratically up and down after completing the ``drink'' action). 
The fourth row presents the greatest challenge, involving four actions. Our method completes at least three, whereas others accomplish only one or two. 
Further comparisons can be found in the \textbf{supplementary video}.
}

\begin{figure}[!t]
    \centering
    \includegraphics[width=1.0\textwidth]{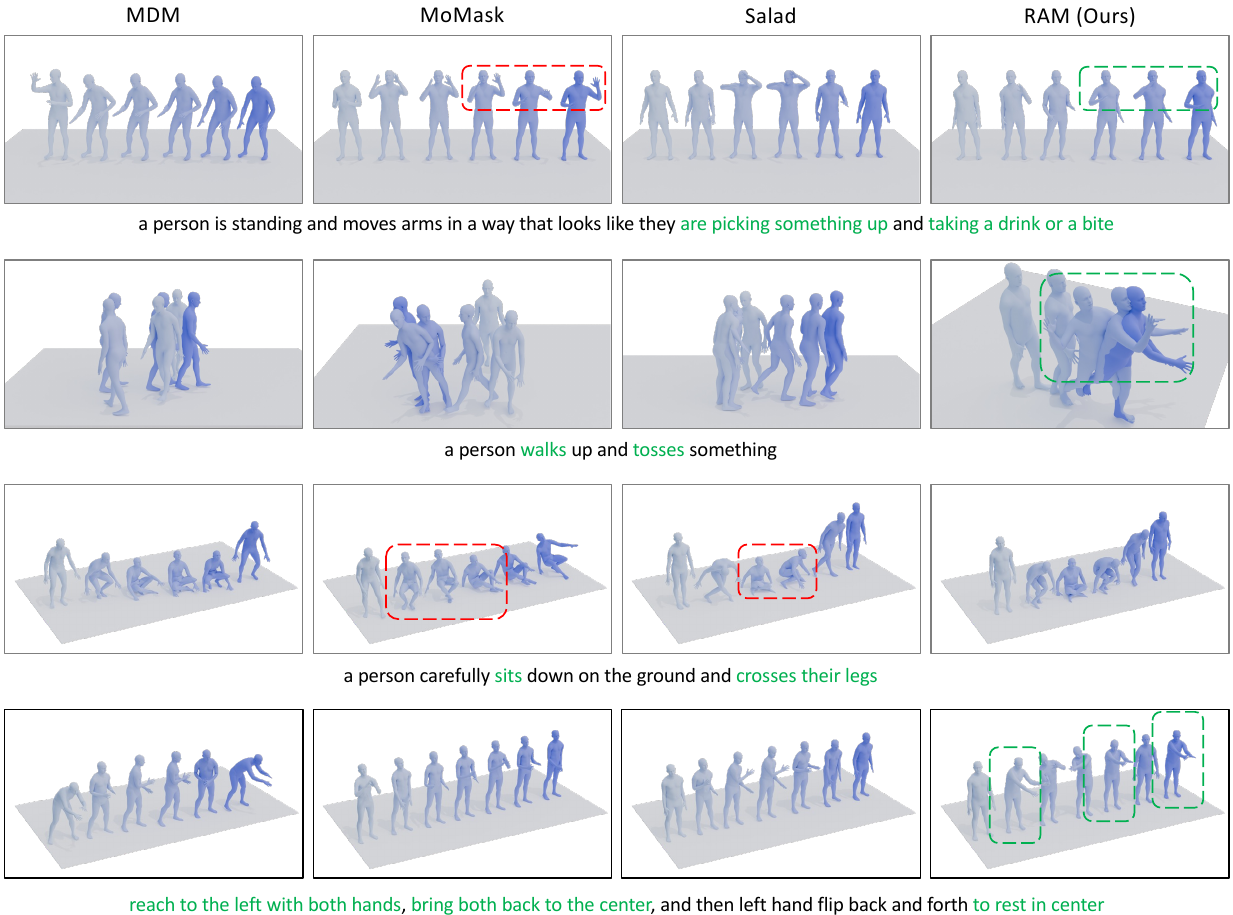}
    \caption{\qualitativeCaption}
    \label{fig:qualitative}
\end{figure}

%% file: Sec/Sec_06_conclusion.tex
\section{Conclusion}

In this work, we present \textbf{\modelname}, a novel framework that leverages motion reconstruction to regularize text-driven motion diffusion models. Our approach focuses on learning a motion-centric latent space via motion reconstruction, specifically designed to capture essential motion dynamics while achieving high semantic resolution. This latent space serves as intermediate supervision for text-to-motion generation, bridging the representational gap between abstract language and high-dimensional, kinematically constrained human motion. We further present Reconstructive Error Guidance, a technique that mitigates error propagation during sampling by exploiting the \highlight{motion} diffusion model's self-correcting ability. Experiments show that \modelname achieves state-of-the-art performance on standard benchmarks. 
One possible limitation of \modelname is that we adopt a single token to represent the entire textual description. In the future, we plan to try finer-grained conditions that contain multiple tokens for long-horizon motion generation.

\medskip

\noindent
{\qheading{Acknowledgments.} This work was supported by the National Natural Science Foundation of China under Grants 62476099 and 62076101, Guangdong Basic and Applied Basic Research Foundation under Grants 2024B1515020082 and 2023A1515010007, and the TCL Young Scholars Program.}

%% file: Sec/Sec_07_sup.tex
\input{Fig/Fig_03_userstudy}

\section{User Study}
\label{sup:user}
To further validate the effectiveness of our method from a perceptual perspective, we conduct a user study comparing three methods: MoMask~\cite{guo2024momask}, Salad~\cite{hong2025salad}, and our proposed \modelname. We randomly select 30 text prompts from the test set and generate one motion sample per method for each prompt, resulting in 90 motion samples in total.

We recruit 15 participants, each holding at least a bachelor's degree. Each participant is asked to evaluate all 30 text prompts. For each prompt, the participant views the three generated motions (presented in random order to avoid bias) and ranks them from best to worst based on two criteria: (1) text-motion alignment, i.e., how well the generated motion matches the semantic content of the text, and (2) motion quality, i.e., the naturalness and realism of the generated motion. Participants are instructed to consider both criteria comprehensively when providing their rankings.

The results of the user study are presented in \cref{fig:user}, which shows the distribution of rankings across all evaluations. Our method achieves the best overall performance, securing the first rank in over 50\% of the cases. This result demonstrates that our approach generates motions that are not only better aligned with the input text but also exhibit superior quality compared to MoMask and Salad.

\section{Inference Efficiency}
\label{sup:effi}
\input{Tab/Tab_08_efficiency}

The proposed Reconstructive Error Guidance (REG) introduces an additional reconstruction step for previous predictions during inference, which increases inference time. However, experiments show that \modelname requires substantially fewer inference steps than most commonly used motion diffusion models. As discussed in \cref{sec4.2}, training uses $T=50$ diffusion steps. For efficient inference, we automatically select 20 denoising steps by linearly spacing them within the interval $[0, \ldots, T-1]$, resulting in the following indices: 
\begin{align*} t = [& 49,\ 46,\ 44,\ 41,\ 39,\ 36,\ 34,\ 31,\ 28,\ 26,\ 23,\ 21,\ 18,\ 15,\ 13,\ 10,\ 8,\ 5,\ 3,\ 0]. \end{align*}
We compare MDM (50steps) and several configurations where REG is applied at different steps, reporting both their average inference time per sentence (AITS) \cite{chen2023executing} and the resulting generation quality. AITS is calculated on the HumanML3D test set by setting the batch size to 1 and excluding model and dataset loading time. Note that MDM (50steps) is an improved variant of the original MDM, offering higher inference efficiency and better generation results compared to those reported in the original paper \cite{shafir2023human}.

The results summarized in \cref{tab:efficiency} show that, due to fewer inference steps, \modelname achieves significantly higher efficiency even when REG is enabled at every step (except the initial step  , which lacks a previous prediction). Notably, enabling REG only in the early denoising steps already leads to a marked improvement in FID. This supports our claim in the introduction that early denoising steps—responsible for recovering motion from nearly pure noise—are particularly prone to generating error patterns, and thus benefit most from the application of REG.

\highlight{To further characterize the overhead, we measure the per-step cost of each guidance configuration. At a batch size of 32, the per-step time ratio of \textit{baseline}:\textit{CFG}:\textit{REG}:\textit{CFG+REG} is approximately $1{:}1.95{:}2.18{:}3.13$, i.e.\ REG is essentially at parity with CFG per step. Combined with the early-window analysis above, applying REG only to the first 6 steps adds just $26\%$ inference time (AITS $0.226{\to}0.284$\,s) while improving FID by $71\%$ ($0.132{\to}0.038$), offering a favorable quality–efficiency trade-off.}

\section{Hyperparameter Analysis}
\label{sup:hyper}

In this section, we analyze the impact of key hyperparameters to verify our design choices and understand their influence on model performance. Our investigation covers three main aspects: (1) \textit{Internal loss parameters}—evaluating $\beta$ and $\tau$, where $\beta$ controls how strongly the latent alignment loss $L_\text{latent}$ updates the motion encoder $E_m$, and $\tau$ shapes the geometry of the motion latent space through the sharpness of similarities in self-regularization; (2) \textit{Loss weights}—assessing the balance between motion-centric latent alignment and self-regularization; and (3) \textit{Latent dimension}—exploring the trade-off between representation capacity and model compactness.

\input{Tab/Tab_04_loss}

\paragraph{Internal Loss Parameters.}\cref{tab:loss} illustrates how the internal parameters $\beta$ and $\tau$ influence generation quality. $\beta$ controls the gradient flow from the latent alignment loss $L_\text{latent}$ to the motion encoder $E_m$, with $\beta$ equal to $0$ fully blocking the gradient and $\beta$ equal to $1$ allowing unrestricted gradient flow. Our results show that allowing equal proximity between text and motion latents (that is, when $\beta$ is equal to $1$) is suboptimal, as this alignment comes at the expense of motion information in the motion latents. In fact, reducing the gradient flow to $E_m$ improves performance, with the best results achieved when $\beta$ is set to $0.01$. We attribute this to the constant evolution of the motion latent space during training, which increases the difficulty of latent alignment. Moreover, enforcing strong alignment gradients on $E_m$ can interfere with learning a motion-discriminative latent space while it is still evolving, making the alignment objective harder to satisfy. By setting $\beta$ to $0.01$, we ease the alignment process while preserving essential motion information in the latent space.

Another parameter $\tau$ determines the sharpness of similarity in $L_\text{sr}$. A smaller $\tau$ produces sharper similarities, pushing motion latents farther apart; if too extreme, this can distort the structure of the motion manifold. In contrast, a larger $\tau$ smooths the similarity, relaxing the constraints between latents, but may reduce gains in semantic resolution. Empirically, we found $\tau=1$ offers a desirable balance.

\input{Tab/Tab_07_weight}

\paragraph{Loss Weights.}We investigate the impact of the weights $w_\text{latent}$ and $w_\text{sr}$, which correspond to motion-centric latent alignment and self-regularization, respectively. The results are presented in \cref{tab:weight}. It can be observed that setting either weight to zero results in a significant performance drop. However, as long as both weights are nonzero, changes in their values have only a minor effect on performance. These findings highlight the importance of each objective component and demonstrate \modelname's robustness to variations in weight assignment.

\input{Tab/Tab_06_dim}

\paragraph{Encoder Latent Dimension.}The dimensionality of the encoder's latent space, $d_E$—which determines the size of the motion and text latents—is a key hyperparameter in our model. A larger $d_E$ can capture more intricate details but increases the model size and may raise the risk of overfitting, while a smaller $d_E$ may result in information loss. In our main experiments, we set $d_E$ to 256. Here, we further explore how varying $d_E$ influences \modelname's performance.

As shown in \cref{tab:dim}, altering $d_E$ leads to only minor fluctuations in performance, indicating that \modelname is relatively robust to this hyperparameter. Interestingly, even when $d_E$ is halved to 128, \modelname's performance only decreases slightly. This suggests that the learned latent space is highly compact.

\section{Implementation Details of Latent Space Training Strategies}
\label{sup:latent_comparison_details}

This section provides the detailed implementation configurations for the comprehensive comparison of latent space training strategies discussed in \cref{sec:4.4abl}.

\paragraph{Network Architecture and Hyperparameters.} 
To ensure a strictly fair comparison, all evaluated variants (Variants A--E and our proposed method) share identical foundational configurations, differing only in their training objectives applied within the latent space. Specifically, all models employ the exact same diffusion-based decoder and maintain equivalent model architectures with the same number of parameters. Furthermore, the optimization hyperparameters, including the learning rate, batch size, and the total number of training steps, are kept strictly consistent across all experiments.

\paragraph{Training Objectives.} 
Regarding the loss functions, for any loss weights that are not explicitly specified in the variant descriptions in the main text, the default value is set to 1.0. Additionally, since our framework does not employ a Variational Autoencoder (VAE) structure, none of the compared variants utilize the Kullback-Leibler (KL) divergence loss. The specific loss weights for bidirectional latent alignment and cross-modal contrastive learning (\eg, temperature $\tau=0.1$) follow the exact settings described in the main paper to faithfully reproduce the representative methods (such as TEMOS and TMR).

\paragraph{Inference Settings.}
At inference time, to isolate the impact of the latent space representations on generation quality, all methods apply standard classifier-free guidance (CFG). To ensure fairness in evaluating the baseline capabilities of the latent spaces, REG is not employed for any of the variants during this specific evaluation.

\input{Fig/Fig_04_tsne}

\paragraph{Visualizing the Motion-Centric Latent.}
\highlight{To directly verify that \modelname learns a motion-centric latent space, we apply joint t-SNE to paired text and motion latents on the HumanML3D test set, as shown in \cref{fig:tsne}. The two-stream baseline (Variant~A) forms compact text clusters but a diffuse motion distribution, revealing a modality gap. The contrastive variant (Variant~E, i.e.\ TMR \cite{petrovich2023tmr}) aligns the two distributions and exposes cluster-level structure. \modelname yields the sharpest clusters with the tightest within-cluster interleaving of paired text and motion latents, corroborated by the per-panel pairwise cosine similarity. This provides direct evidence that our motion-centric alignment preserves motion structure while pulling the text condition onto the motion manifold.}

\section{Leave-One-Out Ablation}
\label{sup:loo}
\input{Tab/Tab_10_leaveoneout}

\highlight{To complement the incremental ablation in \cref{tab:ablation}, we additionally report a leave-one-out study in \cref{tab:loo}, where each component is removed in isolation from the full \modelname. Removing $L_{\text{latent}}$ causes the most severe degradation (FID $0.032{\to}0.424$), as the text and motion latents are no longer aligned. Removing $L_{\text{sr}}$ degrades FID to $0.064$ and Top-1 R-Precision to $0.536$, confirming that self-regularization is essential for a discriminative motion latent. Removing REG raises FID to $0.132$ while leaving R-Precision unchanged, consistent with REG primarily improving motion realism rather than semantic alignment. None of the reduced variants approaches the full model, demonstrating that the components are complementary rather than redundant.}

\input{Fig/Fig_05_reg}

\section{Empirical Analysis of REG's Correction}
\highlight{\cref{fig:reg}(a) examines REG at a single denoising step ($B{=}128$). Viewing the text condition as an anchor in the motion latent space (\cref{sec:obj}), the distance from the prior reconstruction latent to the text anchor serves as a proxy for the magnitude of the previous step's error. We find that this distance correlates positively with the magnitude of the REG correction (median Pearson $r{\approx}0.62$): the more erroneous the previous prediction, the stronger the correction REG applies. REG thus self-scales to where the model is wrong, rather than applying a uniform push.}

%% file: Fig/Fig_03_userstudy.tex
\newcommand{\userCaption}{
User study results on the HumanML3D test set comparing RAM with state-of-the-art methods. We conducted a perceptual study to evaluate human preferences based on a holistic assessment of two key dimensions: motion quality and semantic alignment. Participants were instructed to rank the generated motions from different methods by jointly considering these factors. The results indicate that RAM achieves the best overall performance, demonstrating its superior capability in generating motions that are both realistic and semantically accurate.
}

\begin{figure}
    \centering
    \includegraphics[width=0.8\linewidth]{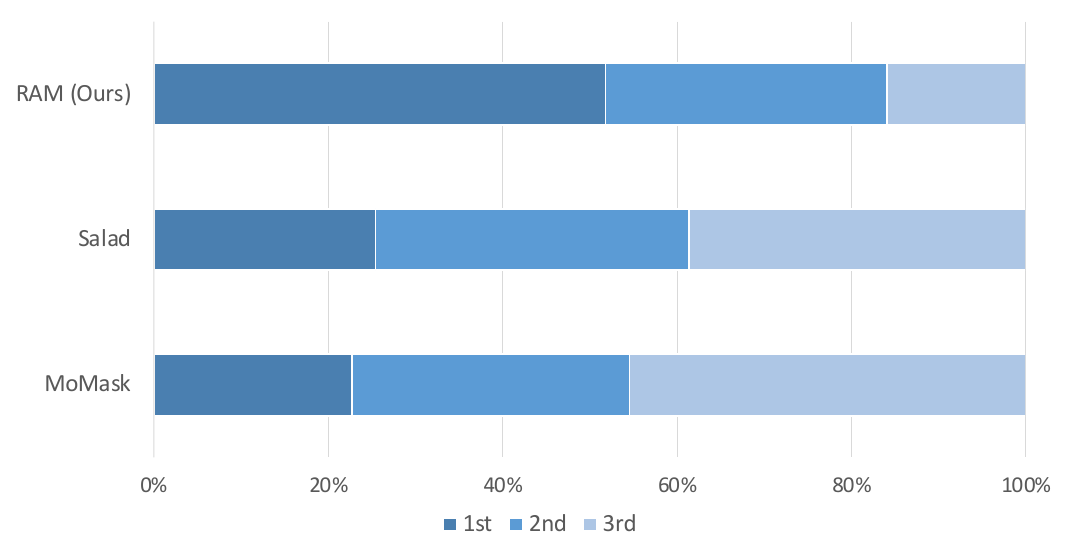}
    \caption{\userCaption}
    \label{fig:user}

\end{figure}

%% file: Tab/Tab_08_efficiency.tex
\begin{table*}
\centering
\setlength{\tabcolsep}{3pt}
\scriptsize
\caption{Experiments on inference efficiency. We compare the Average Inference Time per Sentence (AITS) and generation quality of RAM (20 steps) against MDM (50 steps) with varying REG configurations. The results demonstrate that RAM achieves higher efficiency (lower AITS) than MDM even when REG is fully enabled. Moreover, applying REG only in the early denoising steps yields significant improvements in FID, confirming that correcting errors early in the diffusion process is critical for mitigating error propagation and enhancing motion realism.}
\begin{tabular*}{\textwidth}{@{\extracolsep{\fill}}cc|c|c|ccc}
\toprule
\multirow{2}{*}{Method} & \multirow{2}{*}{enable REG at step $t$} & \multirow{2}{*}{AITS$\downarrow$} & \multirow{2}{*}{FID$\downarrow$} & \multicolumn{3}{c}{R-Precision}\\
 & & &  & Top 1 & Top 2 & Top 3 \\
\midrule
\multirow{2}{*}{\shortstack{MDM\\(50steps)}} &
\multirow{2}{*}{None} &
\multirow{2}{*}{$0.490$} &
\multirow{2}{*}{$0.398^{\pm .010}$} &
\multirow{2}{*}{$0.456^{\pm .002}$} &
\multirow{2}{*}{$0.646^{\pm .003}$} &
\multirow{2}{*}{$0.752^{\pm .002}$} \\ 
& & & & & & \\
\hline
\multirow{5}{*}{\shortstack{\modelname\\(20steps)}}
 & None & $0.226$ & $0.132^{\pm .005}$ & $0.561^{\pm .002}$ & $0.752^{\pm .002}$ & $0.838^{\pm .002}$ \\
& [46,44] & $0.235$ & $0.075^{\pm .003}$ & $0.561^{\pm .002}$ & $0.750^{\pm .002}$ & $0.838^{\pm .001}$ \\
& [46, 44, 41, 39] & $0.261$ & $0.046^{\pm .002}$ & $0.561^{\pm .003}$ & $0.750^{\pm .002}$ & $0.838^{\pm .002}$ \\
& [46, 44, 41, 39, 36, 34] & $0.284$ & $0.038^{\pm .002}$ & $0.560^{\pm .003}$ & $0.750^{\pm .001}$ & $0.838^{\pm .002}$ \\
& All except the inital step & $0.398$ & ${0.032^{\pm .002}}$ & ${0.561}^{\pm .003}$ & ${0.751}^{\pm .002}$ & ${0.839}^{\pm .002}$ \\
\bottomrule
\end{tabular*}

\label{tab:efficiency}
\end{table*}

%% file: Tab/Tab_04_loss.tex
\begin{table}[t]
\centering
\setlength{\tabcolsep}{6pt}
\scriptsize
\caption{Effect of loss parameters $\beta$ and $\tau$. We study the gradient control parameter $\beta$ in latent alignment and the temperature $\tau$ in self-regularization. A small $\beta$ (e.g., 0.01) yields the best performance by preserving essential motion information, while unrestricted flow ($\beta=1$) degrades results. For $\tau$, we find that the model is relatively robust to this parameter, with $\tau=1.0$ achieving the best performance.}
\begin{tabular*}{\linewidth}{@{\extracolsep{\fill}}cc|c|ccc|cc}
\toprule
\multirow{2}{*}{$\beta$} & \multirow{2}{*}{$\tau$} & \multirow{2}{*}{FID$\downarrow$} & \multicolumn{3}{c|}{R-Precision$\uparrow$} & \multirow{2}{*}{MM D$\downarrow$} & \multirow{2}{*}{Diversity}\\
 & & & Top 1 & Top 2 & Top 3 & & \\
\midrule
1.00 & 1.0 & $0.278^{\pm .008}$ & $0.498^{\pm .002}$ & $0.696^{\pm .002}$ & $0.797^{\pm .002}$ & $3.001^{\pm .008}$ & $9.777^{\pm .075}$ \\
0.10 & 1.0 & $0.132^{\pm .003}$ & $0.530^{\pm .001}$ & $0.728^{\pm .003}$ & $0.823^{\pm .003}$ & $2.847^{\pm .007}$ & $9.669^{\pm .067}$ \\
0.01 & 1.0 & $0.032^{\pm .002}$ & $0.561^{\pm .003}$ & $0.751^{\pm .002}$ & $0.839^{\pm .002}$ & $2.716^{\pm .007}$ & $9.487^{\pm .084}$ \\
0.00 & 1.0 & $0.045^{\pm .003}$ & $0.563^{\pm .002}$ & $0.751^{\pm .003}$ & $0.839^{\pm .002}$ & $2.707^{\pm .006}$ & $9.438^{\pm .053}$ \\ \hline
0.01 & 2.0 & $0.055^{\pm .003}$ & $0.561^{\pm .002}$ & $0.753^{\pm .003}$ & $0.842^{\pm .002}$ & $2.704^{\pm .007}$ & $9.537^{\pm .078}$ \\
0.01 & 1.0 & $0.032^{\pm .002}$ & $0.561^{\pm .003}$ & $0.751^{\pm .002}$ & $0.839^{\pm .002}$ & $2.716^{\pm .007}$ & $9.487^{\pm .084}$ \\
0.01 & 0.5 & $0.039^{\pm .003}$ & $0.560^{\pm .002}$ & $0.751^{\pm .002}$ & $0.841^{\pm .002}$ & $2.721^{\pm .008}$ & $9.533^{\pm .076}$ \\
\bottomrule
\end{tabular*}

\label{tab:loss}
\end{table}

%% file: Tab/Tab_07_weight.tex
\begin{table*}
\centering
\setlength{\tabcolsep}{6pt}
\scriptsize
\caption{Effect of loss weights. We evaluate the impact of varying the weights for the self-regularization ($w_{sr}$) and latent alignment ($w_{latent}$) objectives. The quantitative results show that RAM maintains stable performance across different configurations, indicating that the model is robust to variations in these hyperparameters.}
\begin{tabular*}{\textwidth}{@{\extracolsep{\fill}} cc|c|ccc|cc}
\toprule
\multirow{2}{*}{$w_\text{latent}$} & \multirow{2}{*}{$w_\text{sr}$} & \multirow{2}{*}{FID$\downarrow$} & \multicolumn{3}{c|}{R-Precision$\uparrow$} & \multirow{2}{*}{MM Dist$\downarrow$} & \multirow{2}{*}{Diversity}\\
 & & & Top 1 & Top 2 & Top 3 & & \\
\midrule
1.0 & 1.0 & $0.037^{\pm .002}$ & $0.563^{\pm .003}$ & $0.755^{\pm .002}$ & $0.843^{\pm .002}$ & $2.693^{\pm .008}$ & $9.496^{\pm .094}$ \\
1.0 & 0.5 & $0.055^{\pm .003}$ & $0.560^{\pm .003}$ & $0.751^{\pm .003}$ & $0.841^{\pm .002}$ & $2.703^{\pm .009}$ & $9.482^{\pm .080}$\\
1.0 & 0.1 & $0.063^{\pm .004}$ & $0.555^{\pm .004}$ & $0.747^{\pm .002}$ & $0.837^{\pm .002}$ & $2.729^{\pm .006}$ & $9.532^{\pm .074}$ \\
1.0 & 0.0 & $0.109^{\pm .005}$ & $0.533^{\pm .003}$ & $0.722^{\pm .002}$ & $0.815^{\pm .002}$ & $2.859^{\pm .009}$ & $9.508^{\pm .084}$ \\
0.5 & 1.0 & $0.032^{\pm .002}$ & $0.561^{\pm .003}$ & $0.751^{\pm .002}$ & $0.839^{\pm .002}$ & $2.716^{\pm .007}$ & $9.487^{\pm .084}$ \\
0.1 & 1.0 & $0.056^{\pm .002}$ & $0.534^{\pm .003}$ & $0.730^{\pm .002}$ & $0.822^{\pm .001}$ & $2.839^{\pm .007}$ & $9.465^{\pm .062}$\\
0.0 & 1.0 & $0.422^{\pm .016}$ & $0.476^{\pm .003}$ & $0.673^{\pm .003}$ & $0.778^{\pm .002}$ & $3.134^{\pm .010}$ & $9.451^{\pm .075}$ \\
\bottomrule
\end{tabular*}

\label{tab:weight}
\end{table*}

%% file: Tab/Tab_06_dim.tex
\begin{table*}
\centering
\scriptsize
\setlength{\tabcolsep}{9pt}
\caption{Effect of encoder latent dimension $d_E$. The results indicate that $d_E=256$ yields the optimal balance, achieving the lowest FID score. Furthermore, the model exhibits robustness to dimension variations, with even a reduced dimension of 128 maintaining competitive performance, highlighting the compactness of the learned motion latent space.}
\begin{tabular*}{\linewidth}{@{\extracolsep{\fill}}  c|c|ccc|cc}
\toprule
\multirow{2}{*}{$d_E$} & \multirow{2}{*}{FID$\downarrow$} & \multicolumn{3}{c|}{R-Precision$\uparrow$} & \multirow{2}{*}{MM Dist$\downarrow$} & \multirow{2}{*}{Diversity}\\
 & & Top 1 & Top 2 & Top 3 & & \\
\midrule
128 & $0.043^{\pm .002}$ & $0.546^{\pm .002}$ & $0.739^{\pm .002}$ & $0.829^{\pm .002}$ & $2.767^{\pm .008}$ & $9.509^{\pm .088}$\\
192 & $0.050^{\pm .003}$ & $0.555^{\pm .002}$ & $0.748^{\pm .003}$ & $0.838^{\pm .002}$ & $2.725^{\pm .008}$ & $9.545^{\pm .072}$\\
256 & $0.032^{\pm .002}$ & $0.561^{\pm .003}$ & $0.751^{\pm .002}$ & $0.839^{\pm .002}$ & $2.716^{\pm .007}$ & $9.487^{\pm .084}$ \\
384 & $0.049^{\pm .002}$ & $0.555^{\pm .002}$ & $0.748^{\pm .002}$ & $0.836^{\pm .002}$ & $2.746^{\pm .009}$ & $9.618^{\pm .104}$\\
512 & $0.064^{\pm .003}$ & $0.558^{\pm .003}$ & $0.753^{\pm .002}$ & $0.842^{\pm .002}$ & $2.724^{\pm .007}$ & $9.583^{\pm .085}$\\
\bottomrule
\end{tabular*}

\label{tab:dim}
\end{table*}

%% file: Fig/Fig_04_tsne.tex
\begin{figure}
    \centering
    \includegraphics[width=0.95\linewidth]{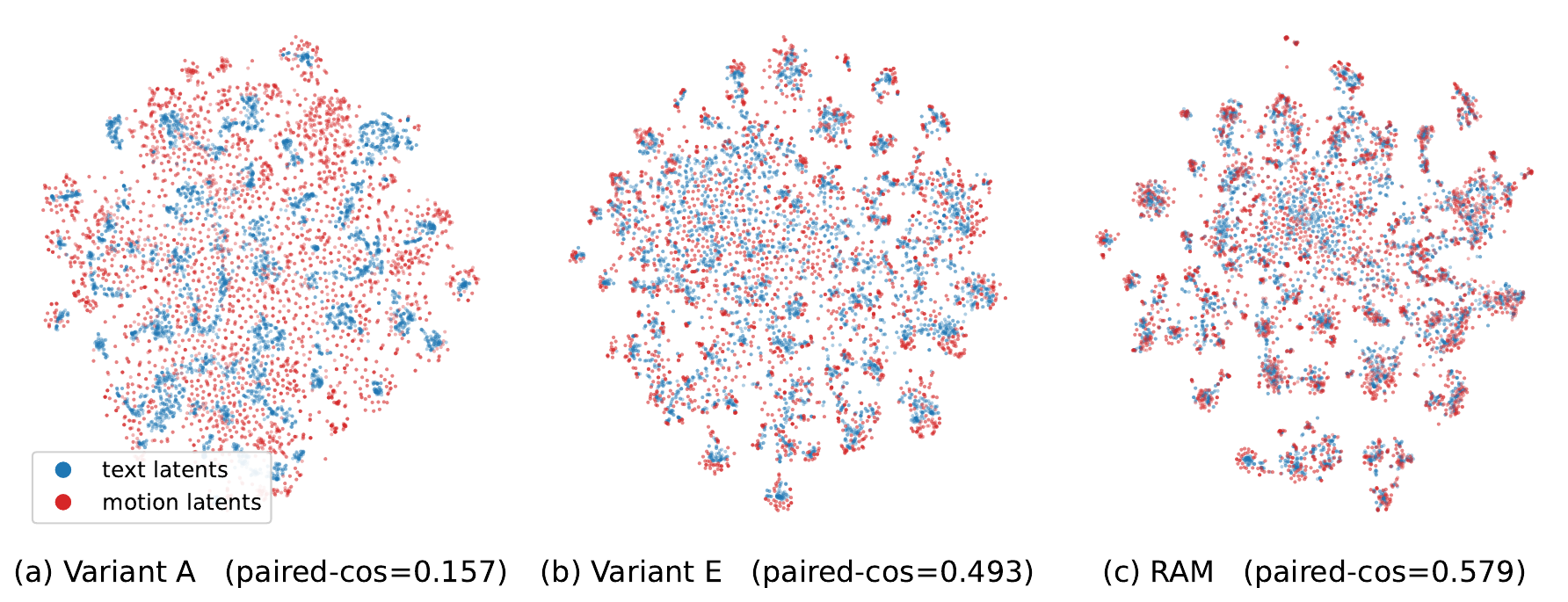}
    \caption{Joint t-SNE of paired text and motion latents on HumanML3D.}
    \label{fig:tsne}

\end{figure}

%% file: Tab/Tab_10_leaveoneout.tex
\begin{table}[t]
\centering
\scriptsize
\caption{Leave-one-out ablation on HumanML3D. Starting from the full \modelname, we remove a single component at a time.}
\begin{tabular}{l|c|c|c|c}
\toprule
 & Full & w/o $L_{\text{latent}}$ & w/o $L_{\text{sr}}$ & w/o REG \\
\midrule
FID$\downarrow$ & $\mathbf{0.032}^{\pm .002}$ & $0.424^{\pm .016}$ & $0.064^{\pm .003}$ & $0.132^{\pm .005}$ \\
R-Precision Top 1$\uparrow$ & $\mathbf{0.561}^{\pm .003}$ & $0.476^{\pm .003}$ & $0.536^{\pm .003}$ & $\mathbf{0.561}^{\pm .002}$ \\
\bottomrule
\end{tabular}
\label{tab:loo}
\end{table}

%% file: Fig/Fig_05_reg.tex
\begin{figure}
    \centering
    \includegraphics[width=0.9\linewidth]{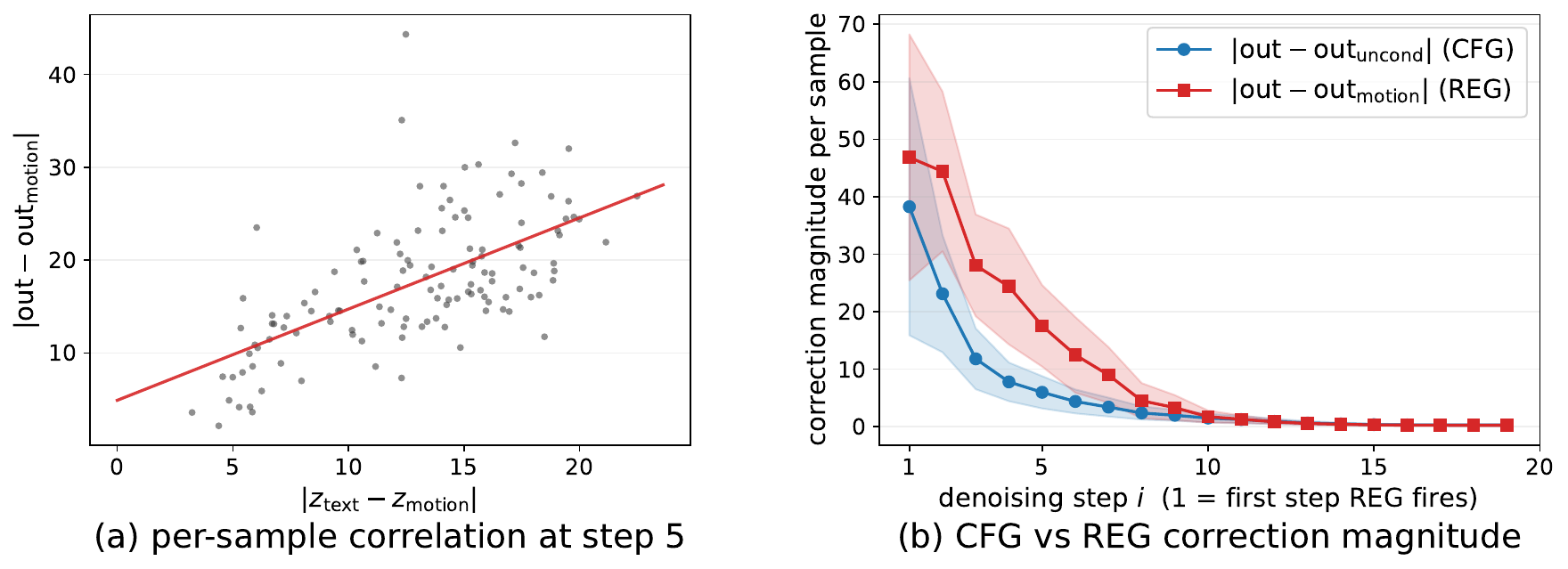}
    \caption{REG: per-sample scatter (a), per-step magnitude (b).}
    \label{fig:reg}
\end{figure}